\documentclass[conference]{IEEEtran}

\usepackage[hyphens]{url}
\usepackage[bookmarks=false, draft]{hyperref}
\hypersetup{breaklinks=true}

\ifCLASSINFOpdf
  \usepackage[pdftex]{graphicx}

\else

\fi

\usepackage{amsmath}

\usepackage{algorithmic}

\usepackage{array}

\ifCLASSOPTIONcompsoc
 \usepackage[caption=false,font=normalsize,labelfont=sf,textfont=sf]{subfig}
\else
 \usepackage[caption=false,font=footnotesize]{subfig}
\fi

\usepackage{booktabs} 
\usepackage{multirow}
\usepackage[strings]{underscore}
\usepackage[english]{babel}

\usepackage{color}
\usepackage{xcolor}

\def\BibTeX{{\rm B\kern-.05em{\sc i\kern-.025em b}\kern-.08em
    T\kern-.1667em\lower.7ex\hbox{E}\kern-.125emX}}

\begin{document}

\title{FR-NAS: Forward-and-Reverse Graph Predictor for Efficient Neural Architecture Search}

\author{\IEEEauthorblockN{Haoming~Zhang and Ran~Cheng\IEEEauthorrefmark{1}}
\IEEEauthorblockA{Department of Computer Science and Engineering, Southern University of Science and Technology, China \\
zhm1998@outlook.com, ranchengcn@gmail.com
}
\IEEEauthorrefmark{1}{\emph{Corresponding author}}
}

\maketitle

\begin{abstract}
Neural Architecture Search (NAS) has emerged as a key tool in identifying optimal configurations of deep neural networks tailored to specific tasks.
However, training and assessing numerous architectures introduces considerable computational overhead. 
One method to mitigating this is through performance predictors, which offer a means to estimate the potential of an architecture without exhaustive training. 
Given that neural architectures fundamentally resemble Directed Acyclic Graphs (DAGs), Graph Neural Networks (GNNs) become an apparent choice for such predictive tasks. 
Nevertheless, the scarcity of training data can impact the precision of GNN-based predictors.
To address this, we introduce a novel GNN predictor for NAS. 
This predictor renders neural architectures into vector representations by combining both the conventional and inverse graph views. 
Additionally, we incorporate a customized training loss within the GNN predictor to ensure efficient utilization of both types of representations.
We subsequently assessed our method  through experiments on benchmark datasets including NAS-Bench-101, NAS-Bench-201, and the DARTS search space, with a training dataset ranging from 50 to 400 samples. 
Benchmarked against leading GNN predictors, the experimental results showcase a significant improvement in prediction accuracy, with a 3\%--16\% increase in Kendall-tau correlation. Source codes are available at https://github.com/EMI-Group/fr-nas.
\end{abstract}



\section{Introduction}

Neural Architecture Search (NAS) plays a pivotal role in the automated generation of high-performing deep neural networks. 
Given a designated dataset, NAS can be viewed as an optimization problem,  which aims to discover an architecture that maximizes accuracy and other performance metrics within a circumscribed search domain.
Typically, NAS algorithms leverage performance metrics from a variety of tested architectures to guide the development of more effective designs. 
Various strategies have been systematically applied to conduct NAS tasks, including Reinforcement Learning~\cite{Tan_2019_CVPR}, Bayesian Optimization~\cite{NEURIPS2018BOTNAS, White_Neiswanger_Savani_2021}, and Evolutionary Algorithms~\cite{lu2019nsga, sun2020evonet}, among others~\cite{Zoph_2018_CVPR,Wu_2019_CVPR}.

However, a neural network's performance is contingent on both its architecture and weight configurations.
Evaluating these architectures demands rigorous training and validation on specific datasets, which often leads to high computational costs, sometimes equivalent to thousands of GPU days~\cite{Zoph_2018_CVPR,pmlr-v97-ying19a}. 
In response to this challenge, recent research has focused on methods to streamline evaluations. 
For example, the weight-sharing method~\cite{cai2020once,chu2021fairnas,tan2023} trains \emph{super networks} which encapsulate all potential architectures, thereby abbreviating the training duration. 
A key advantage of super networks is the ability to inherit weights directly, obviating the need for redundant training. 
 Another group of emerging methods involves the development of \emph{predictors}~\cite{Liu_2018_ECCV, wenwei/NPNAS, NPENAS}. 
Once calibrated on a curated subset of architectures, a predictor is expected to estimate the performance of architectures that have not been empirically tested.

In the realm of NAS predictors, establishing an effective representation is foundational, as it directly influences the quality of the training data. 
Recognizing neural architectures' inherent representation as Directed Acyclic Graphs (DAGs)~\cite{pmlr-v97-ying19a}, it is imperative to harness such topological information. 
Some research initiatives have gravitated towards serializing these DAGs~\cite{sum2020E2EPP,White_Neiswanger_Savani_2021}. 
However, such serialized methods often provide an indirect or even incomplete representation, thus failing to encapsulate the gamut of topological intricacies. 
To bridge this gap, some research efforts have been dedicated to \emph{graph predictors}. 
The Graph Neural Networks (GNNs), including models such as the Graph Convolutional Networks (GCNs)~\cite{kipf2017semisupervised}, Graph Isomorphism Networks (GINs)~\cite{xu2018how}, and Graph Attention Networks (GATs)~\cite{veličković2018graph}, are specifically designed for processing graph-structured data.
By iteratively aggregating neighborhood data, GNNs adeptly distill intricate feature representations for each vertex in a graph. 
This inherent competency enables GNNs to generate robust graph embeddings, which is foundational to performance predictions. 
Empirical studies have shown their superior predictive performance compared to sequence-based predictors~\cite{Chen_2021_ICCV, NPENAS}.

\begin{figure*}[tbp]
\centering
\includegraphics[width=6in]{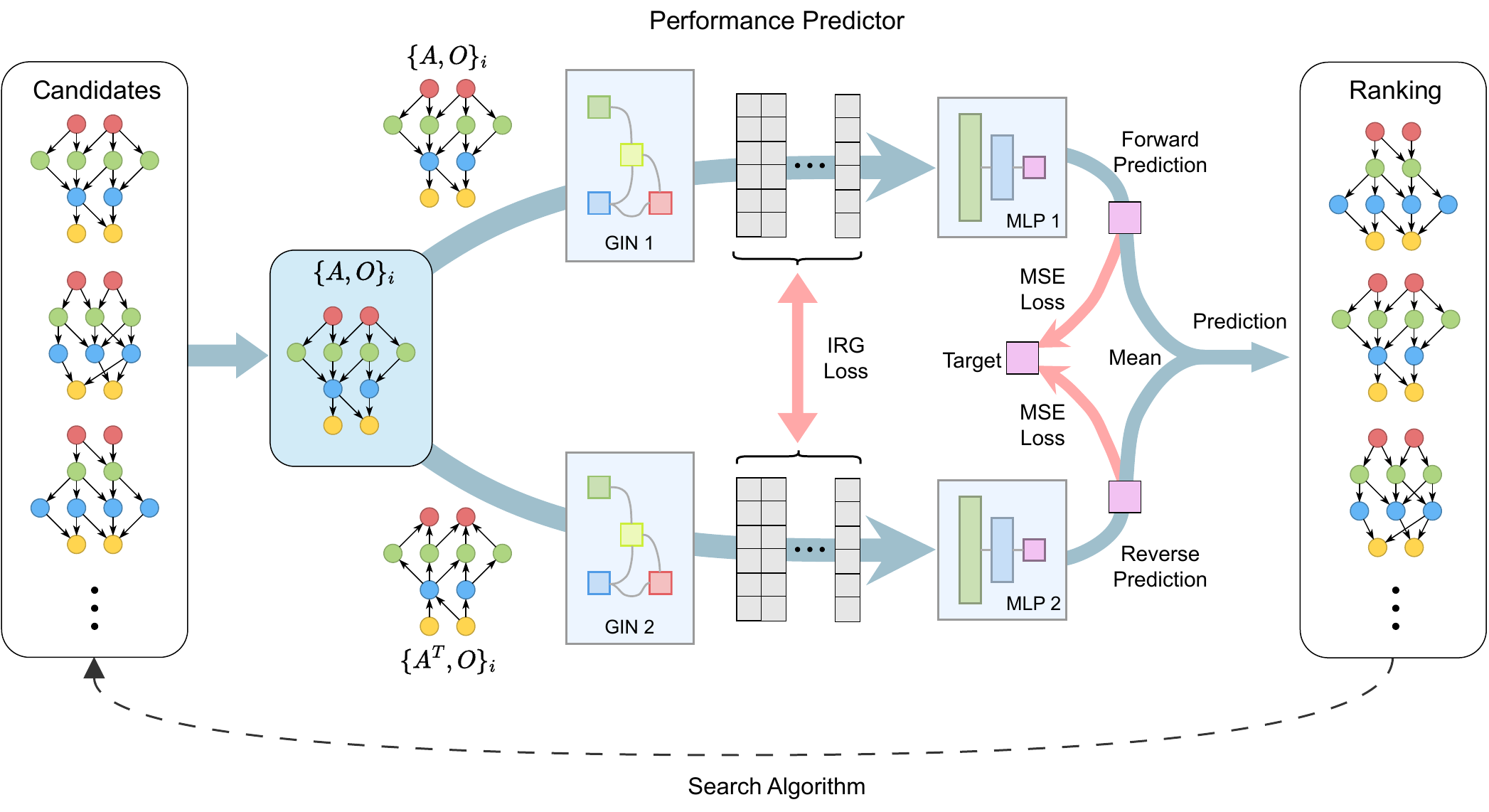}
\vskip -1em
\caption{Framework of our FR-NAS. The architectures are assessed by the performance predictor and represented using both forward and reverse graph encodings (blue arrows). 
These are processed by two GIN encoders into feature vectors, which are then fed to MLPs. 
Two training losses, Instance Relationship Graph (IRG) and Mean Square Error (MSE) losses are incorporated, targeting features and predictions respectively (red arrows).}

\label{fig:nas_framework}
\end{figure*}

While GNN-based predictors have demonstrated their efficacy as performance predictors, most studies use \emph{unidirectional} (forward) representations of the computation graphs during predictor training.
By contrast, the neural architectures are inherently bidirectional, involving both forward and backward propagation phases. 
This raises the question: \emph{Can we harness the inherent bidirectionality of neural architectures to enhance the performance of graph predictors?}
Our research is directed towards answering this question.

We initiate our study by repeatedly training a GNN predictor and analyzing the features extracted by encoders using visualization techniques.
Our observations suggest that in the presence of limited training data, the encoder often faces challenges in effectively embedding features crucial for precise predictions. 
Since valuable features are shared across different representations, employing separate encoders for these variations naturally leads to a collaboration for better data utilization.
To tackle this challenge, we introduce a graph-based predictor that capitalizes on both forward and reverse graph representations, which can be integrated with the general NAS framework (dubbed FR-NAS).
As illustrated in Fig.~\ref{fig:nas_framework},
in the proposed FR-NAS framework, every architecture is depicted as a DAG and its inverse, and fed into distinct GINs. 
To ensure that encoders from varied representations converge towards shared features, we incorporate a feature loss during the training phase at the embedding layer of the GINs.
The main contributions of this work are summarized as follows.

\begin{itemize}
\item Through detailed analysis of the GIN predictor outcomes employing both forward and reverse graph representations, we highlight relationships and distributions of features across various search spaces. Our findings underscore the potential of dual graph representations in enhancing prediction accuracy.
\item We design a performance predictor built upon a new network structure, where both forward and reverse graph depictions of architectures are employed. 
Moreover, we introduce a tailored training loss to ensure congruence in embeddings generated by both GINs.
\item Our comprehensive benchmark experiments compared our method with two leading GNN-based predictors, NPNAS~\cite{wenwei/NPNAS} and NPENAS~\cite{NPENAS}. 
The experimental results, alongside the ablation studies, confirm the effectiveness of our method across various search spaces.
\end{itemize}

\section{Related Work}

This section provides a brief review of the existing literature on performance predictors, techniques utilizing multiple data sources, and contrasts with the domain of contrastive learning.

\subsection{Performance Predictors}

Various strategies have been proposed to encode these graph structured data, with some approaches opting for serialized encodings. 
However, they often struggle with maintaining topological intricacies~\cite{Liu_2018_ECCV}. 
BANANAS~\cite{White_Neiswanger_Savani_2021} suggested a path-based encoding scheme, where the occurrence of all unique paths in DAGs is used as an encoding. However, the length of encoding grows exponentially when facing a larger search space. 
NAO~\cite{NEURIPS2018_933670f1} jointly trained an encoder and a decoder to learn continuous representation along with a performance predictor for gradient optimization. MoSegNAS~\cite{lu2023moseg} utilized ranking loss and large amount of data generated by super network.
Due to GNNs' inherent ability to handle graph data, they become ideal for capturing both topological structures and associated operations. Previous studies have employed GNNs for extracting embeddings from graph-encoded data for subsequent performance prediction. 
For instance, BONAS~\cite{NEURIPS2020_13d4635d} utilized the GCN and added a global node to generate embedding for whole computation graph.
NPNAS~\cite{wenwei/NPNAS} proposed a directed GCN to better handle DAG data, while HOP~\cite{Chen_2021_ICCV} advocated for GATs and applied attention mechanism to aggregate outputs from layers. 
NPENAS~\cite{NPENAS} employed the GIN and isolated node representation, which has demonstrated superiority in handling directed graphs. 
Despite the occasional inclusion of reverse graph data, the general trend seems to favor the development of more powerful predictors primarily oriented around forward graph data.

\subsection{Data Augmentation}
The field of data augmentation continues to innovate, introducing techniques to boost predictor performance. 
HAAP~\cite{Liu_2021_ICCV}, for example, employed homogeneous representations transformed into sequences. This stands in contrast to our method, which maps distinct graph representations to identical labels. 
Methods like OMNI~\cite{NEURIPS2021_ef575e88} combined various data sources, while others like FBNetV3~\cite{Dai_2021_CVPR} and GMAE~\cite{ijcai2022p432} utilized auxiliary information or modified graph data.
D-VAE~\cite{NEURIPS2019_e205ee2a} achieves a comprehensive latent representation for DAGs by merging encoding states from both the graph's forward and reverse message passing scheme. Conversely, SVGe ~\cite{9534092} treats the forward and backward graph encodings independently, jointly decoding both  to reconstruct the respective graph directions distinctly.
However, simply concatenating two data sources may not be sufficient to extract features from DAGs that are relevant for prediction tasks. 
An area yet to be thoroughly explored is the potential of the reverse graph as an alternative data source, a direction which holds promise for producing rich augmentations. 

\subsection{Contrastive Learning}
\label{sec:contrastive}
Contrastive learning has emerged as a prominent technique for producing robust feature embeddings, especially when data is scarce. 
For instance, GraphCL~\cite{NEURIPS2020_3fe23034} presented a graph contrastive framework where perturbation techniques generate variant graphs which are then processed by a GNN encoder. 
In the context of NAS, CTNAS~\cite{Chen_2021_CVPR} assessed the relative performance of diverse architectures using GCNs. 
Another recent approach~\cite{hesslow2021contrastive} leverages an extended data Jacobian matrix for a contrastive network. 
Distinctively, our method employs different graph encodings for the same architectural input.
Unlike existing methods that focus on varying inputs to a single network, ours centers on dual GNN encoders handling different encodings.

\section{Method}
In this section, we present the proposed FR-NAS method as illustrated in Fig.~\ref{fig:nas_framework}.
We begin by outlining the architectural representations input to our predictor, followed by an in-depth exploration of our encoder and predictor designs. 
Then, we introduce the training loss, which is essential for the cohesive training of both the encoder and predictor. Finally, we explain our motivations by conducting empirical analysis.

\subsection{Architecture Encoding}
Since neural architectures can be generally conceptualized as DAGs, we adopt an adjacency matrix $\boldsymbol{A}$ to delineate edges connecting vertices and deploy a sequence of one-hot vectors $\boldsymbol{O}$ to represent operations at each vertex. 
The matrix $\boldsymbol{A} \in \{0,1\}^{N\times N}$ signifies a graph of $N$ vertices, where $\boldsymbol{A}_{i,j}=1$ denotes an edge from vertex $i$ to vertex $j$. 
The one-hot encoding maps operations according to a predefined sequence and generates a consistent-length vector for each vertex. 
Only one index with a value of 1 indicates the operation of the corresponding vertex. 
Specifically, we refer to this type of encoding as \emph{forward} graph encoding, which propagates features in alignment with edge directions, while its \emph{reverse} counterpart is derived by transposing the adjacency matrix to obtain $\boldsymbol{A}^T$.

\subsection{Encoder and Predictor}
We integrate two distinct GINs to process the forward and reverse graph encodings, respectively. 
Specifically, each GIN encoder comprises three consecutive layers, where each layer employs dual fully connected structure with the ReLU activation function.
Subsequently, a Global Mean Pooling (GMP) layer extracts the embedding.

Within a single GIN layer, every vertex aggregates features from its preceding neighbors and its own data. 
This aggregated information then feeds into the ensuing fully connected layers. 
Due to the variant feature trajectory in the reversed graph, sharing weights becomes nonviable. 
Specifically, the encoding procedure of our proposed predictor can be articulated as:
\begin{align}
\boldsymbol{h}_{f} &= \text{Enc}(\boldsymbol{A}, \boldsymbol{O} ; \boldsymbol{W}_1), \\
\boldsymbol{h}_{r} &= \text{Enc}(\boldsymbol{A}^T, \boldsymbol{O} ; \boldsymbol{W}_2),
\end{align}
where $\boldsymbol{A}$ is the adjacency matrix, $\boldsymbol{O}$ is the one-hot encoded operation sequence, $\boldsymbol{h}_f$ and $\boldsymbol{h}_r$ are the features embedded using forward and reverse directed graph data, and $\boldsymbol{W}_1$ and $\boldsymbol{W}_2$ are trainable weights in the encoder.

Following the encoding phase, we design a predictor derived from the embeddings. 
Given potential discrepancies between the initial embeddings from the encoder, deploying a unified predictor to process both embedded features may introduce biases. 
To mitigate this, we utilize two separate fully connected layers, each addressing the feature embeddings from a specific encoder. 
The ensuing prediction is ascertained by averaging the outcomes from the fully connected layers:
\begin{align}
\boldsymbol{p}_{f} = \text{FC}(\boldsymbol{h}_{f}; \boldsymbol{W}_1 ), \\
\boldsymbol{p}_{r} = \text{FC}(\boldsymbol{h}_{r}; \boldsymbol{W}_2 ), \\
\boldsymbol{p} = (\boldsymbol{p}_{f} + \boldsymbol{p}_{r})/2 ,
\end{align}
where $\boldsymbol{p}_{f}$ and $\boldsymbol{p}_{r}$ are the prediction results using forward and reverse graph data, $\text{FC}$ is a sequence of two fully connected layers, and $\boldsymbol{p}$ is the final output of the predictor.

\subsection{Training Loss}

Our foremost training goal is to collaboratively refine both encoders with an emphasis on creating a harmonized representation strategy that promotes accurate predictions.
With sparse training data, however, the encoders may not optimally capture pivotal features that significantly impact prediction accuracy (as will be discussed in Section~\ref{sec:2.4}).
To mitigate this limitation, we propose a mutual learning strategy where the encoders reciprocally reinforce their ability to identify and exploit shared features.

In this regard, our training loss is meticulously formulated to bridge and diminish the discrepancies in the embedded features emanating from each encoder. 
However, a challenge arises due to the inherent variability in the ordering of elements within embedding vectors, resulting an element-wise comparison of encoder outputs infeasible. Moreover, encoders with different topologies as input follow different information propagation mechanisms.
Given such challenges, relying on the MSE loss for the output embedding vectors can be counterproductive.


It is observed that neural architectures with close performances often share common features. In contrast, architectures with diverging performances usually have distinct features. The complex relationships within these architectures' representations from both encoders provide critical insights. 
Such pairing relationships remain robust against random ordering of features. Building on these insights and drawing inspiration from the IRG~\cite{Liu_2019_CVPR} framework, we develop a loss function aimed at reducing the discrepancy between the two encoders, defined as
\begin{align}
\mathcal{L}_{e} &= \frac{1}{M^2} \sum_{i=1}^{M}\sum_{j=1}^{M} \left( \Vert \boldsymbol{h}_{fi}-\boldsymbol{h}_{fj}\Vert_2^2 - \Vert \boldsymbol{h}_{ri}-\boldsymbol{h}_{rj}\Vert_2^2 \right) ^2 ,
\end{align}
where \(M\) represents the number of architectures in one batch, \(\boldsymbol{h}_{fi}\) and \(\boldsymbol{h}_{ri}\) symbolize features of the \(i\)-th architecture embedding emanating from the forward and reverse graph encoders, respectively.
Notably, unlike the original IRG which learns existing knowledge from a teacher, our loss is designed to bidirectionally minimize the difference from scratch between the two encoders.

Subsequently, we employ the MSE of predictions and true performances as the prediction loss. The losses are calculated for the forward predictor ($\mathcal{L}_{pf}$) and reverse predictor ($\mathcal{L}_{pr}$) respectively:
\begin{align}
\mathcal{L}_{pf} &= \frac{1}{M} \sum_{i=1}^{M}(\boldsymbol{p}_{fi}-\boldsymbol{y}_i)^2, \\
\mathcal{L}_{pr} &= \frac{1}{M} \sum_{i=1}^{M}(\boldsymbol{p}_{ri}-\boldsymbol{y}_i)^2,
\end{align}
where \(\boldsymbol{p}_{fi}\) and \(\boldsymbol{p}_{ri}\) indicate the predicted performance of the \(i\)-th architecture, and \(\boldsymbol{y}_i\) signifies the actual performance of the \(i\)-th architecture. 
Both \(\mathcal{L}_{pf}\) and \(\mathcal{L}_{pr}\) can be directly minimized as there is no interdependence between the two predictor components.

In summary, the entire training loss is formulated as:
\begin{align}
\mathcal{L}_{1} &= (1-\lambda)\mathcal{L}_{pf} + \lambda \mathcal{L}_{e}, \\
\mathcal{L}_{2} &= (1-\lambda)\mathcal{L}_{pr} + \lambda \mathcal{L}_{e}, 
\end{align}
where \(\lambda\) is a weight coefficient balancing the impact of the two losses. 
Given that the gradients are intertwined due to multiple losses, we train the predictor iteratively. 
On each iteration, \(\mathcal{L}_1\) and \(\mathcal{L}_2\) are determined and backpropagated sequentially to update the weights.

\subsection{Empirical Analysis}
\label{sec:2.4}

In this subsection, we focus on the training and evaluation of a straightforward graph-based predictor and delve into a detailed examination of the IRG matrix produced from the output embedding vectors of the encoders. 
Our analysis reveals the disadvantages of relying solely on one type of representation. 
This reliance often results in suboptimal embedding performance. 
In addition, we discuss the potential benefits of using two distinct representations to extract more detailed features.

\begin{figure*}[tb]
\centering
\subfloat[50 samples]{
\includegraphics[width=2in]{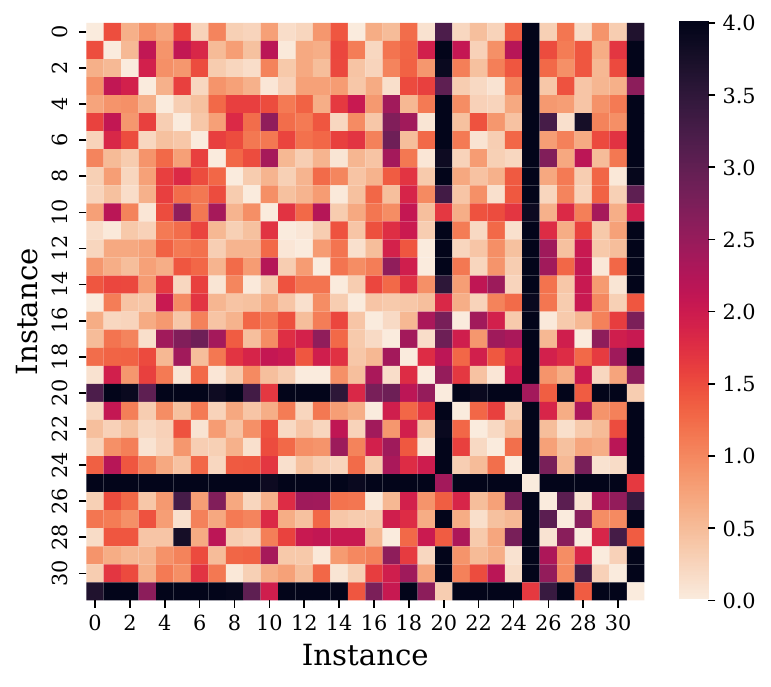}
}
\hfil
\subfloat[200 samples]{
\includegraphics[width=2in]{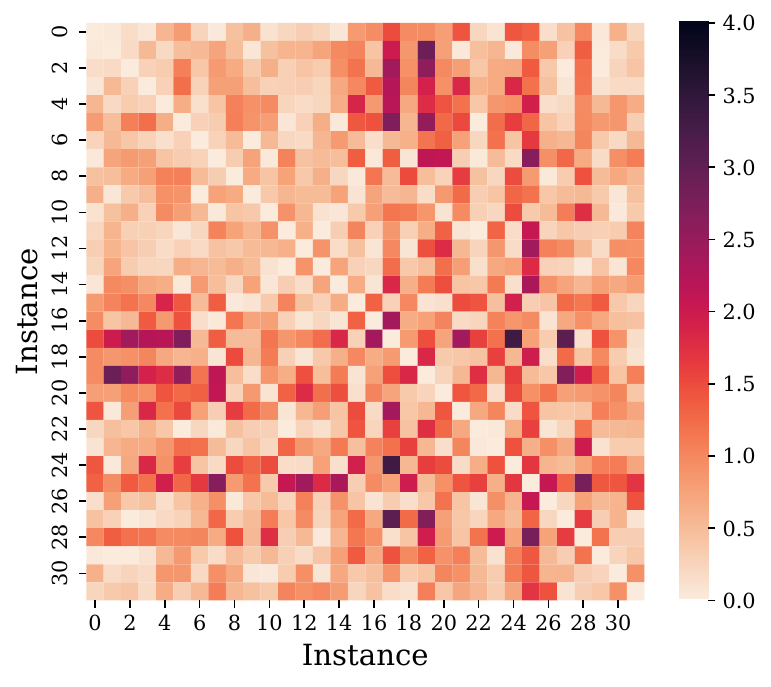}
}
\hfil
\subfloat[400 samples]{
\includegraphics[width=2in]{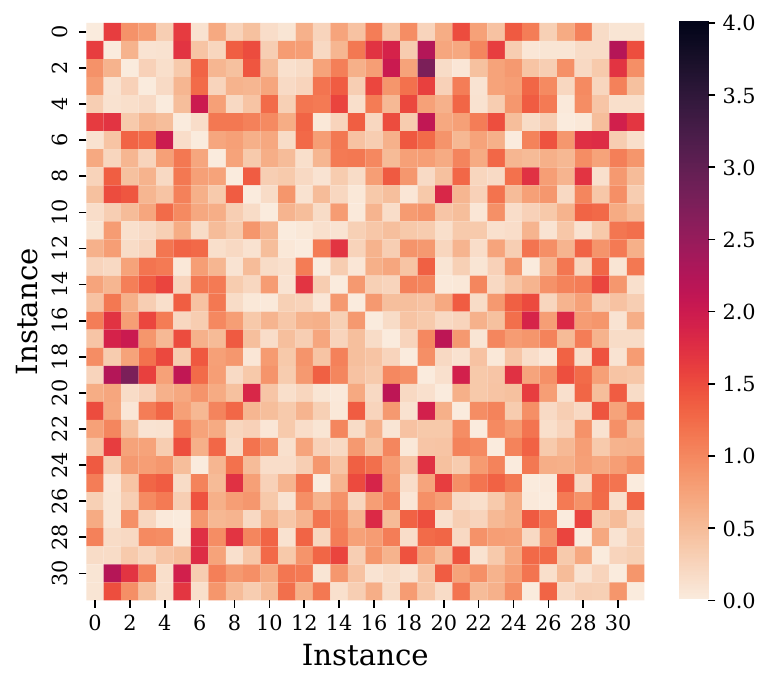}
}
\caption{Differences in the IRG matrix of embedding vectors when trained without the proposed feature loss, using 50, 200, and 400 samples from NAS-Bench-201, respectively.}
\label{fig:Empirical_Analysis}
\end{figure*}

\begin{figure*}[tb]
\centering
\subfloat[50 samples with feature loss]{
\includegraphics[width=2in]{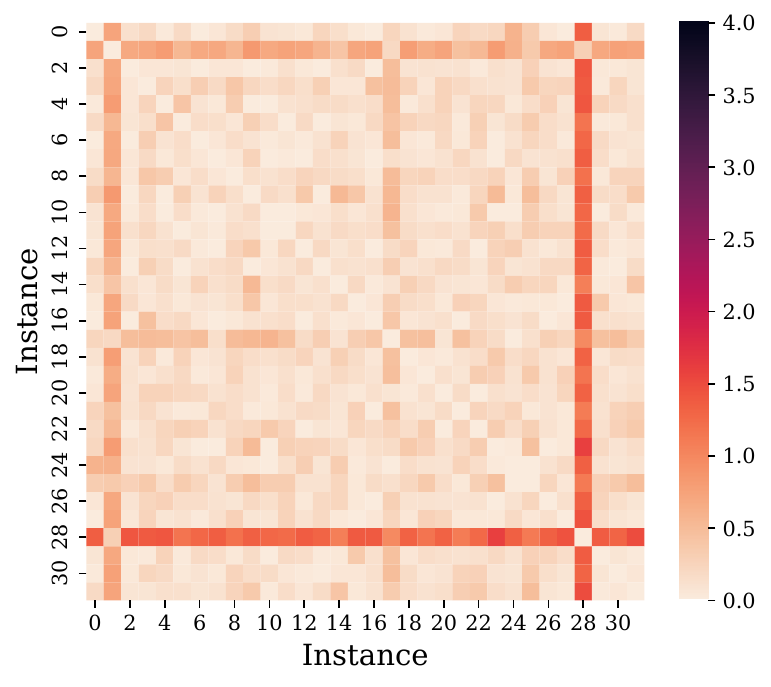}
}
\hfil
\subfloat[50 samples w/o feature loss]{
\includegraphics[width=2in]{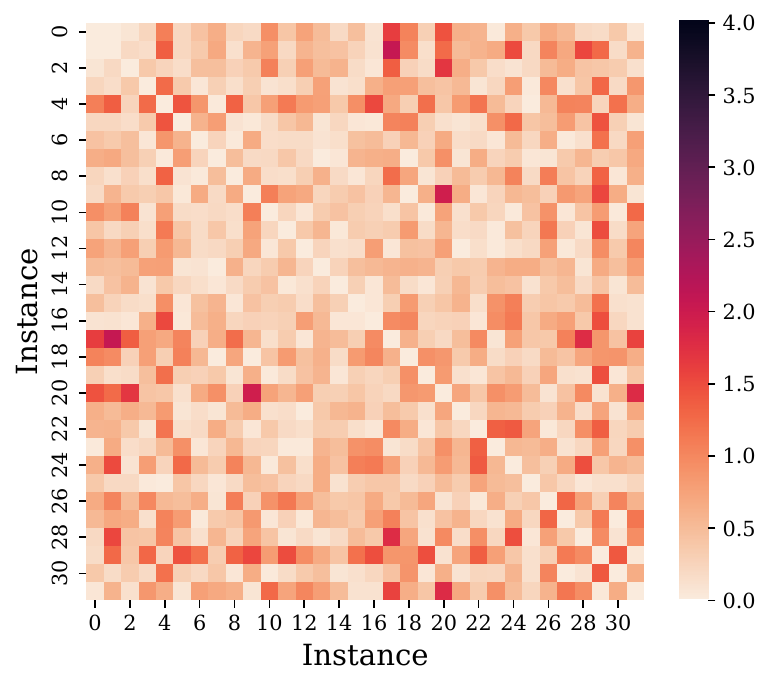}
}
\hfil
\subfloat[400 samples w/o feature loss]{
\includegraphics[width=2in]{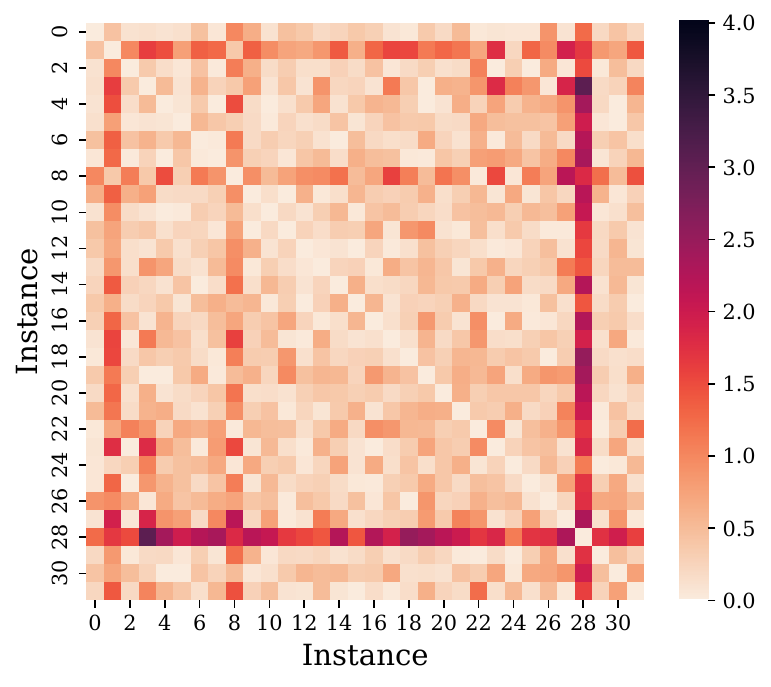}
}
\caption{Differences in the IRG matrix of embedding vectors {when trained with(out) the proposed feature loss, using 50 and 400 samples from DARTS search space, respectively}. }
\label{fig:Empirical_Analysis_2}
\end{figure*}

We employed the GIN predictor on two distinct search spaces: the simpler NAS-Bench-201~\cite{dong2020bench} and the more complex DARTS~\cite{liu2018darts}. 
After randomly sampling 400 architectures, we selected 200 architectures out of them, from which we further selected 50.
These divisions allow us to study the impact of increasing the volume of training data. 
Each predictor was trained from scratch on these subsets. 
Following this, we computed the IRG matrix for an additional set of 32 unseen architectures. 
From this matrix, we derived the element-wise distance matrix for both the encoders and decoders. 
The results are showcased in Fig.~\ref{fig:Empirical_Analysis} and Fig.~\ref{fig:Empirical_Analysis_2}, where each data point on a heatmap represents the absolute difference in the distances encoded by the two separate encoders:
\begin{align}
\text{Diff}(i,j) = \bigl| \Vert \boldsymbol{h}_{fi}-\boldsymbol{h}_{fj}\Vert_2^2 - \Vert \boldsymbol{h}_{ri}-\boldsymbol{h}_{rj}\Vert_2^2 \bigr|. 
\end{align}

As indicated in Fig.~\ref{fig:Empirical_Analysis}(a), utilizing only 50 architectures for training results in significant differences between the two encoders. 
Given the constrained search space of NAS-Bench-201, expanding the training dataset reduces these discrepancies, as shown in Fig.~\ref{fig:Empirical_Analysis}(b).
We also ran several trials with different subsets of the training data. 
We found a consistent pattern: In 42 out of 50 trials, there was a noticeable reduction in IRG loss as the sample size increased. 
This observation strengthens our belief that with adequate training data, both encoders tend to align their knowledge. 
This alignment is particularly evident since they both perform identical regression tasks. 
However, in scenarios with limited data, the lack of comprehensive information affects both encoders. 
Combined with differing topologies, this can lead to changes in how information travels within the GNN layers. 
Consequently, the two GNN encoders might produce varied embeddings for the same architectures. 
Given that high-quality predictors often yield similar IRG matrices for subsequent predictions, a smart approach would be to synchronize the embeddings of both predictors. 

The results in Figs.~\ref{fig:Empirical_Analysis_2} were obtained from the more challenging DARTS search space.
For Fig.~\ref{fig:Empirical_Analysis_2}(a), we applied the feature loss method introduced earlier but with a smaller data size. 
On the other hand, Figs.~\ref{fig:Empirical_Analysis_2}(b) and ~\ref{fig:Empirical_Analysis_2}(c) show outcomes when trained with same or larger data size, respectively, but without using the feature loss method. 
The 1st and 28th instances stand out as they represent unseen architectures with rare and high prediction errors. 
Predicting these unfamiliar architectures accurately is challenging, thus leading to disparities in the encoded distances compared to other instances. 
When dealing with limited training data, the omission of a feature loss often leads to the neglect of intricate details. 
By contrast, our proposed method successfully captures these nuanced features, notwithstanding the limited size of the training dataset.

\section{Experiments}
In this section, we assess the performance of our method on three benchmark search spaces: NAS-Bench-101~\cite{pmlr-v97-ying19a}, NAS-Bench-201~\cite{dong2020bench}, and the DARTS search space.
First, we introduce the settings of the search space, baseline algorithms, and hyperparameters. 
Then, we present the overall performance results on different training data sizes and search spaces. 
Finally, we perform a comprehensive ablation study to demonstrate the effectiveness of our method.

\subsection{Experiment Setup}

\subsubsection{Search Spaces}  
Operations are represented on nodes, which can be directly represented using our architecture representation. 
The NAS-Bench-101 search space is a cell-based space consisting of around 423k unique convolutional architectures. 
These architectures are trained and tested on CIFAR-10~\cite{krizhevsky2009learning}, providing ample labeled data for predictor training. 
Three types of operations are applied to operation vertices. 
Unlike NAS-Bench-101, the original representation for NAS-Bench-201 and DARTS represents operations on edges and gathers features on nodes. 
We modified the representation such that operations are on vertices. 
Additionally, since the DARTS search space comprises approximately \(10^{21}\) architectures, we directly sampled from the training data rather than the proxy data of predictors from NAS-Bench-301\cite{zela2022surrogate}. 
In this experiment, we used validation accuracy on CIFAR-10 as our training data.
For all labeled data, we converted the validation accuracy into an error percentage. 
All experiments were conducted using datasets provided by EvoXBench~\cite{lu2023evoxbench}.

\subsubsection{Baselines} 
We evaluated our proposed FR-NAS in comparison with two state-of-the-art GNN-based peer methods: NPENAS~\cite{NPENAS} and NPNAS~\cite{wenwei/NPNAS}. 
We also compared the representative algorithms mentioned above using different model-based predictors, namely NAO~\cite{NEURIPS2018_933670f1}, BONAS~\cite{NEURIPS2020_13d4635d}, and BANANAS~\cite{White_Neiswanger_Savani_2021} , which are implemented by NASLib~\cite{mehta2022bench}.

NPNAS employs GCN as a predictor and introduces a directed GCN layer to address the limitations of GCN on directed graphs. 
This layer comprises two GCN layers: one taking forward input and the other reverse input. 
The output features of both inputs are then averaged to produce the final output. 
Since it integrates both forward and reverse graphs in a single layer, we cannot apply our framework to such a GCN encoder. 
We compared the results of the NPNAS predictor with our proposed FR-NAS predictor to demonstrate the advantages of utilizing two representations. 
The GIN predictor from NPENAS highlights the capabilities of GIN layers in architecture representation on directed graphs and has shown superior performance. 
For fair comparisons, we employed the same data preprocessing method as in the original NPENAS.

\subsubsection{Performance Metric} 
We used the Kendall rank correlation coefficient \(\tau\) as the metric. 
As NAS aims to identify architectures with superior performance, ranking accuracy is crucial for performance prediction. 
Higher ranking accuracy increases the likelihood of the NAS algorithm finding optimal architectures. 

\begin{figure}[!b]
\centering
\includegraphics[width=3in]{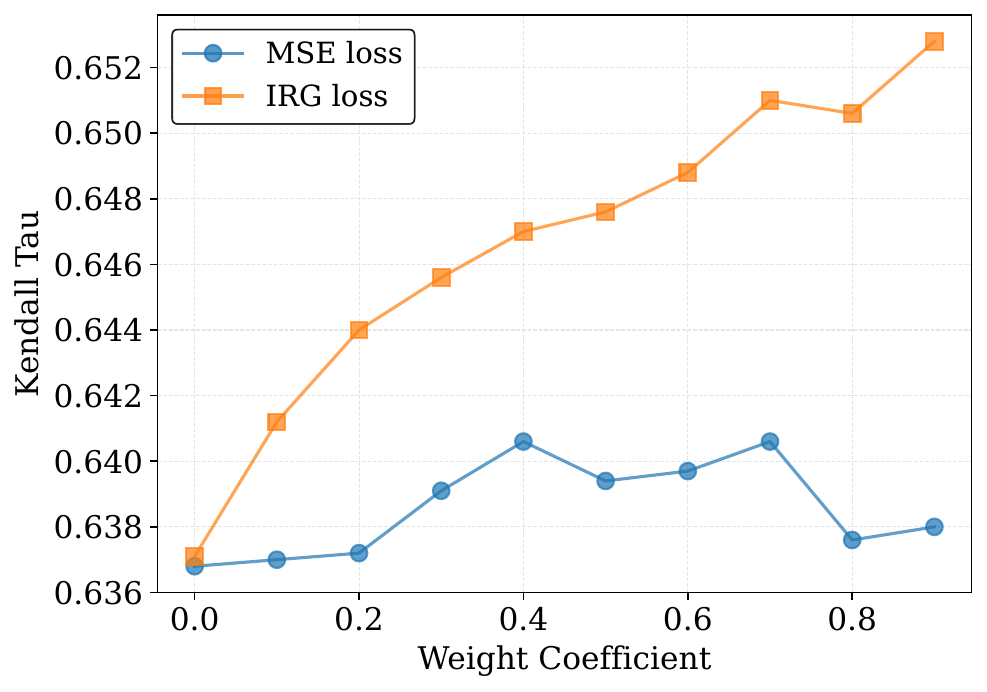}
\caption{Parameter sensitivity analysis of the weight coefficient \(\lambda\).}
\label{fig:parameter_sensitivity_analysis}
\end{figure}

\subsubsection{Hyperparameter Setting}
The only additional parameter in our method is the weight coefficient \(\lambda\), which was fixed at 0.8 according to our parameter sensitivity analysis in Fig.~\ref{fig:parameter_sensitivity_analysis}.  
For fair comparisons, we used consistent hyperparameter and  structure of the two GIN predictors as in the single GIN predictor of NPENAS. 
We also implemented the GCN predictor using settings consistent with NPNAS. 
The hidden size of the two fully connected layers within the GIN layer is set to 32, except for the first GIN layer. 
This first layer is designed to handle the one-hot feature, which has dimensions of 6 for NAS-Bench-101, 7 for NAS-Bench-201 and 13 for DARTS, while the prediction layer employs a hidden size of 16. 
When training the GIN predictor for both NPENAS and FR-NAS, batch normalization is added after each GIN block and fully connected layer. 
A dropout layer, with a dropout rate of 0.1, is placed after the GIN encoder's embedding. 

For optimization, we used the Adam optimizer~\cite{kingma2017adam} with a cosine annealing learning rate, where the initial learning rate and weight decay were set at $5 \times 10^{-3}$ and $1 \times 10^{-4}$, respectively. 
We used data sets ranging in size from 50 to 400 for training, while testing was conducted with a data set of 5,000 samples. 
Training of the GIN predictor underwent for 300 epochs with a batch size of 16. 
To reduce potential biases from sampling, 200 independent random experiments were conducted.

\begin{table*}[t]
\centering
\caption{Comparison of Kendall $\tau$ correlation for predictors trained with varying data sizes across NAS-Bench-101, NAS-Bench-201, and DARTS search spaces. Values represent the mean Kendall $\tau$ over 200 runs, with the best result in each section highlighted in bold.}

\begin{center}
\begin{tabular}{llrrrrrr}
\toprule
\multirow{2}{*}{\bf Search Space} & \multirow{2}{*}{\bf Algorithm}  & \multicolumn{6}{c}{\bf Training Data Size} \\ \cmidrule{3-8} 
    &      & \multicolumn{1}{c}{50} & \multicolumn{1}{c}{100} & \multicolumn{1}{c}{150} & \multicolumn{1}{c}{200} & \multicolumn{1}{c}{300}  & \multicolumn{1}{c}{400}   \\ \midrule
\multirow{6}{*}{\shortstack{NAS-Bench-101}} & NAO & 0.2653 & 0.2932 & 0.3306 & 0.3506 & 0.3984 & 0.4346 \\
& BONAS & 0.2630 & 0.3076 & 0.3196 & 0.3332 & 0.3754 & 0.3847 \\
& BANANAS & 0.5077 & 0.6102 & 0.6331 & 0.6691 & 0.7151 & 0.7330 \\
& NPNAS & 0.5499 & 0.5636 & 0.5812 & 0.5924 & 0.6138 & 0.6309 \\
& NPENAS  & 0.4470 & 0.5692 & 0.6196 & 0.6444 & 0.6755 & 0.6942 \\
& FR-NAS (ours)   & \bf 0.5556 & \bf 0.6596 & \bf 0.6950 & \bf 0.7131 & \bf 0.7359 & \bf 0.7496 \\ 
\midrule
\multirow{6}{*}{\shortstack{NAS-Bench-201}} & NAO & 0.5208 & 0.5321 & 0.5363 & 0.5498 & 0.5520 & 0.5559 \\
& BONAS & 0.3790 & 0.3959 & 0.4302 & 0.4801 & 0.4870 & 0.4941 \\
& BANANAS & 0.2050 & 0.3241 & 0.3679 & 0.4185 & 0.4797 & 0.5211 \\
& NPNAS & 0.4810 & 0.5221 & 0.5538 & 0.5865 & 0.6202 & 0.6400 \\
& NPENAS & 0.5587 & 0.6560 & 0.6983 & 0.7266 & 0.7631 & 0.7882 \\
& FR-NAS (ours)  & \bf 0.6305 & \bf 0.7129 & \bf 0.7476 & \bf 0.7730 & \bf 0.8061 & \bf 0.8262 \\
\midrule
\multirow{6}{*}{\shortstack{DARTS}} & NAO & 0.2843 & 0.3020 & 0.2844 & 0.3172 & 0.3387 & 0.3302 \\
& BONAS & 0.2764 & 0.3148 & 0.3448 & 0.3756 & 0.3980 & 0.4001 \\
& BANANAS & 0.0363 & 0.0447 & 0.0378 & 0.0455 & 0.0520 & 0.0595 \\
& NPNAS & 0.4813 & 0.5568 & 0.5934 & 0.6068  & 0.6254 & 0.6389 \\
& NPENAS & 0.4656 & 0.5409 & 0.5691 & 0.5851 & 0.6125 & 0.6275 \\
& FR-NAS (ours)  & \bf 0.5334 & \bf 0.5989 & \bf 0.6237 & \bf 0.6425 & \bf 0.6666 & \bf 0.6818 \\ 
\bottomrule
\end{tabular}
\end{center}
\label{table:KT-pred-compare}

\end{table*}

\subsection{Results}

The experimental results are summarized in Table~\ref{table:KT-pred-compare}. 
Generally, our method outperforms peer methods in all cases. 
Some of the methods from NASLib may not perform well as they are not designed for larger search spaces or such evaluation metrics. 
For instance, the path encoding from BANANAS is longer in DARTS search space.
The GIN predictor from NPENAS excels with larger data sizes. 
While the GIN predictor employs an MLP to aggregate representations, it can assimilate more information.
By contrast, the GCN predictor outperforms the GIN predictor with a training data size of 50 in the larger search spaces, indicating a substantial performance decline for the GIN predictor. 
Using the same GIN encoder settings, our method enhances performance, especially with smaller data sizes.

\subsection{Ablation Study}
There are two main components in the FR-NAS predictor for performance improvements:
(1) the two GIN encoders taking forward and reverse graph as input; (2) the feature loss combined with prediction loss. 

To evaluate the effectiveness of the two mechanisms, we compared FR-NAS with some variants of NPENAS for an ablation study: a pair of predictors using the same GIN architecture but taking only the forward direction of graph encoding as inputs (denoted as NPENAS-Forward), a forward-and-reverse pair using the same input and architecture as our method (denoted as NPENAS-FR). 

\begin{figure*}[!h]
\centering
\subfloat[NAS-Bench-101]{
\includegraphics[width=2.25in]{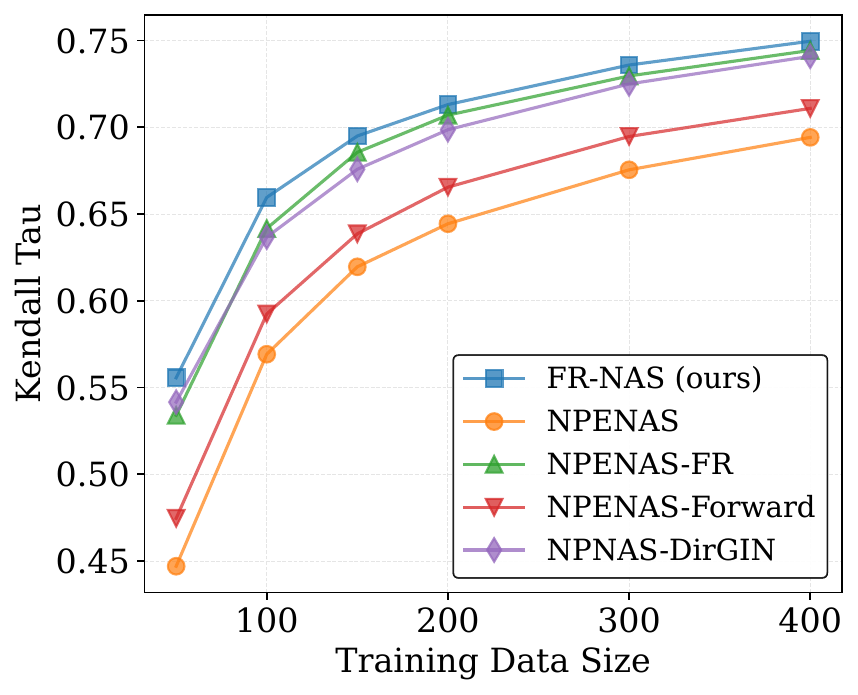}
}\hfill
\subfloat[NAS-Bench-201]{
\includegraphics[width=2.25in]{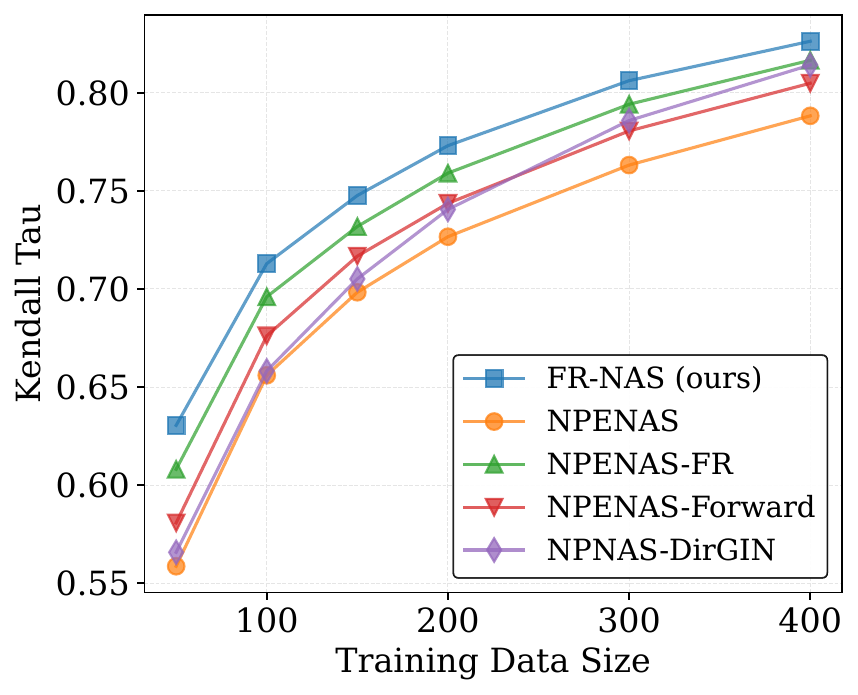}
}\hfill
\subfloat[DARTS]{
\includegraphics[width=2.25in]{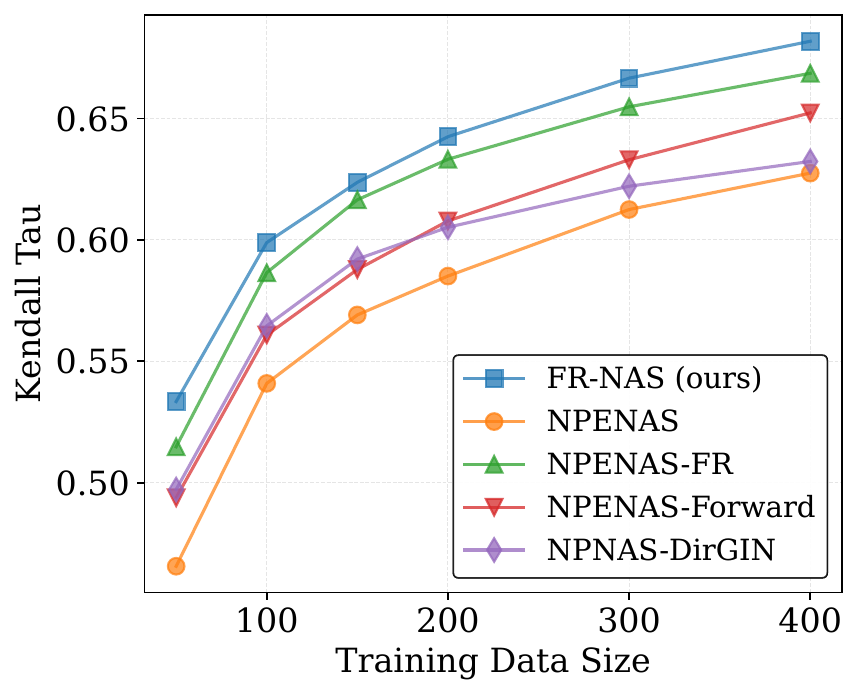}
}
\caption{Comparison of variants on three benchmark datasets. NPENAS-Forward: employing the same architecture as FR-NAS but taking only the forward direction of graph encoding as inputs. NPENAS-FR: using the same inputs and architecture as FR-NAS but without using  our training method. NPNAS-DirGIN: a GIN adaptation of NPNAS featuring bidirectional GIN layers.}
\label{fig:ablation_study}
\end{figure*}

As shown by the results in Fig.~\ref{fig:ablation_study}, a simple forward and reverse paired predictor can be better than a pair of single direction predictors. 
Moreover, when applying the new training loss from our method, the performance is generally improved, especially when the training data size is small. 
This indicates that such a structure can better utilize two representations.

To further demonstrate the efficacy of our method, we constructed a directed GIN predictor based on the GCN variant as introduced by NPNAS, which we term as NPNAS-DirGIN. 
It is noteworthy that NPNAS inherently considers both the forward and reverse graph configurations on a layer-wise basis. 
A comparative analysis presented in Table~\ref{table:KT-pred-compare} indicates that our method yields significantly improved outcomes. However, such comparison raises an ambiguity. 
Given that distinct GNN layers are applied for our method and NPNAS, it remains uncertain whether both representations can be efficiently harnessed by the mechanism of NPNAS.
To address this, we designed a directed GIN layer that amalgamates the outputs of two GIN layers: one emphasizing the forward propagation of information and the other focusing on the reverse. 
Preliminary results for NPNAS-DirGIN in Figs.~\ref{fig:ablation_study}(a) and ~\ref{fig:ablation_study}(b) suggest that this combined layer aligns most closely with the outcomes from individual forward and reverse ensembles within smaller search realms, while it is evident that the mechanisms adopted by NPNAS might not align seamlessly with GIN layers. 
This incongruity potentially arises since there is no dedicated mechanism ensuring that both facets of a directed GIN layer assimilate meaningful features during training. 
In scenarios involving expansive searches, such as those using DARTS as depicted in Fig.~\ref{fig:ablation_study}(c), the likelihood of deriving additional insights from dual representations appears to be reduced.

\section{Conclusion}
In this research, we introduce an improved GNN predictor that simultaneously leverages forward and reverse graph representations. 
By delving into the intricacies of GNN predictors and understanding the impact of various graph representations on encoding and prediction, we identify the merit of bidirectional topological consideration. 
This insight suggests potential enhancements in prediction accuracy. 
The adapted training loss, which is tailored to optimize feature extraction, further supports this hypothesis. 
Experimental results not only validate our method but also highlight a performance improvement over traditional GNN predictors. 
Our research provides strategies to amplify the effectiveness of GNN predictors, particularly in data-limited settings.

\bibliographystyle{IEEEtran}
\bibliography{mybib}

\end{document}